\documentclass[pdflatex,sn-mathphys,Numbered]{sn-jnl}

\usepackage{graphicx}
\usepackage{multirow}
\usepackage{amsmath,amssymb,amsfonts}
\usepackage{amsthm}
\usepackage{mathrsfs}
\usepackage[title]{appendix}
\usepackage{xcolor}
\usepackage{textcomp}
\usepackage{manyfoot}
\usepackage{booktabs}
\usepackage{algorithm}
\usepackage{algorithmicx}
\usepackage{algpseudocode}
\usepackage{listings}
\usepackage{bm}
\usepackage{subcaption}

\theoremstyle{thmstyleone}

\theoremstyle{thmstyletwo}

\theoremstyle{thmstylethree}

\def\eg{\text{e.g.}} 
\def\ie{\text{i.e.}} 
 
\def\etc{\text{etc}}

\usepackage[most]{tcolorbox}
\tcbuselibrary{breakable}
\newtcolorbox{mybox}[1]{
    colback=white,
    colframe=black,
    title=#1,
    fonttitle=\bfseries,
    boxed title size=title,
    enhanced,
    attach boxed title to top left={yshift=-6.5mm},
    sharp corners,
    coltitle=black,
    boxed title style={sharp corners, colframe=black, colback=orange!20, frame hidden},
    breakable
}
\usepackage{upgreek}

\begin{document}

\title[Article Title]{Chemical knowledge-informed framework for privacy-aware retrosynthesis learning}

\author[1]{\fnm{Guikun} \sur{Chen}}\email{guikunchen@gmail.com}

\author[1]{\fnm{Xu} \sur{Zhang}}\email{xu.zhang@zju.edu.cn}

\author[2]{\fnm{Xiaolin} \sur{Hu}}\email{xiaolinhu@ruc.edu.cn}

\author[2]{\fnm{Yong} \sur{Liu}}\email{liuyonggsai@ruc.edu.cn}

\author[1]{\fnm{Yi} \sur{Yang}}\email{yangyics@zju.edu.cn}

\author*[1]{\fnm{Wenguan} \sur{Wang}}\email{wenguanwang.ai@gmail.com}

\affil[1]{\orgdiv{College of Computer Science and Technology}, \orgname{ Zhejiang University}, \orgaddress{\city{Hangzhou}, \postcode{310058}, \state{Zhejiang}, \country{China}}}

\affil[2]{\orgdiv{Gaoling School of Artificial Intelligence}, \orgname{Renmin University of China}, \orgaddress{\city{Beijing}, \postcode{100872}, \country{China}}}

\abstract{Chemical reaction data is a pivotal asset, driving advances in competitive fields such as pharmaceuticals, materials science, and industrial chemistry. Its proprietary nature renders it sensitive, as it often includes confidential insights and competitive advantages organizations strive to protect. However, in contrast to this need for confidentiality, the current standard training paradigm for machine learning-based retrosynthesis gathers reaction data from multiple sources into one single edge to train prediction models. This paradigm poses considerable privacy risks as it necessitates broad data availability across organizational boundaries and frequent data transmission between entities, potentially exposing proprietary information to unauthorized access or interception during storage and transfer. In the present study, we introduce the chemical knowledge-informed framework (CKIF), a privacy-preserving approach for learning retrosynthesis models. CKIF enables distributed training across multiple chemical organizations without compromising the confidentiality of proprietary reaction data. Instead of gathering raw reaction data, CKIF learns retrosynthesis models through iterative, chemical knowledge-informed aggregation of model parameters. In particular, the chemical properties of predicted reactants are leveraged to quantitatively assess the observable behaviors of individual models, which in turn determines the adaptive weights used for model aggregation. On a variety of reaction datasets, CKIF outperforms several strong baselines by a clear margin.

}


\maketitle

\section{Introduction}\label{sec:intro}

Retrosynthesis is a fundamental technique in organic chemistry that involves designing synthetic routes for a target molecule by working backwards from the desired product to commercially available starting materials. It is of great importance as it allows chemists to discover novel reactions for scarce or even brand-new molecules, optimize existing synthesis pathways, and circumvent processes that are costly, risky, and time-consuming~\cite{subbaraman2011flawed,struble2020current}.

There are three main types of methods for learning retrosynthesis models. Template-based methods rely on reaction templates that specify how a target molecule can be transformed into reactants by applying certain rules~\cite{wang2025towards}. Reaction templates can be either manually encoded by experts \cite{corey1985computer} or automatically extracted from large reaction databases \cite{segler2017neural,coley2017computer,dai2019retrosynthesis,baylon2019enhancing,chen2021deep}. Template-based methods use subgraph isomorphism to match a target molecule with a suitable template and then apply the template rules to generate reactants. This process can be repeated until feasible synthetic routes are found. Template-free methods, implemented in an end-to-end fashion, treat retrosynthesis as a sequence to sequence problem and use generative models to directly generate a sequence of reactants from a target molecule without using any templates or rules. The sequence can be represented by different formats, such as simplified molecular input line entry system (SMILES) tokens \cite{weininger1988smiles,tetko2020state,lin2020automatic,kim2021valid,wan2022retroformer,zhong2022root} or graph edit actions \cite{seo2021gta,tu2022permutation,zhong2023retrosynthesis}, depending on the representation of molecules. Semi-template methods combine template-based and template-free methods by dividing retrosynthesis into two steps: reaction center identification and synthon completion \cite{sun2021towards}. Reaction center identification aims to find the atoms or bonds that are involved in the reaction, while synthon completion is devised to add the necessary atoms or groups to complete the reactants. Semi-template methods use machine learning (ML) models to predict the reaction centers and then use templates or other methods to complete the synthons \cite{shi2020graph,yan2020retroxpert,somnath2021learning,wang2021retroprime}. In a nutshell, all the existing methods follow the same paradigm --- gathering raw reaction data and training prediction models on it.

This reliance on data, however, presents significant challenges to developing ML models for retrosynthesis. Chief among these is the necessity for extensive chemical reaction datasets~\cite{zhang2024retrosynthesis}, the compilation of which entails considerable expense. This expense arises from the requirement for specialized equipment, skilled labor, and material resources, as well as the conversion of unstructured records into structured data. While a limited number of chemical reaction datasets are available open source~\cite{lowe2012extraction,Lowe2017}, the majority are proprietary, maintained by commercial entities~\cite{goodman2009computer,lawson2014making,mayfield2017pistachio}. Furthermore, companies may develop new chemical reactions crucial for advancing key fields (\eg, materials~\cite{bozbag2012synthesis,fayette2014chemical}, drugs~\cite{blakemore2018organic}, and energy solutions~\cite{binder2009simple,poizot2011clean}), but are often hesitant to share this data. This proprietary stance is primarily due to two reasons. On the one hand, reaction data is often sensitive and confidential, as it may reveal proprietary or classified information that belongs to a specific entity or organization~\cite{shimizu2015privacy,simm2021splitting}. For example, a pharmaceutical company may want to protect its findings that leads to the synthesis of a new drug, or a government agency may want to safeguard its reaction data that relates to their future development. On the other hand, reaction data is often valuable and competitive~\cite{muetterties1977molecular,schooler2011unpublished}, which may confer an advantage or disadvantage to a certain entity or organization. For instance, a rival company may want to gain insight into the developed reactions of its competitors to gain an edge in the market, or a hostile state may want to access the frequently used reactions of its adversary to prepare for competitions in advance. 

As aforementioned, only a limited number of retrosynthesis datasets are available openly, and sharing chemical reaction data with external parties might expose sensitive information (products, processes, intellectual property, \etc.), which may cause compromised interests and unfair competition. This often leads to the emergence of ``data islands'' where data collected in different organizations remains siloed and inaccessible to the broader research community. Such fragmentation impedes progress significantly, as the collective benefit of shared insights is lost, slowing the pace of scientific discovery and application. Therefore, there is a pressing need to develop frameworks for privacy-preserving retrosynthesis learning that enable the sharing of valuable knowledge without compromising proprietary information. Developing such frameworks would facilitate a collaborative environment conducive to the advancement of retrosynthesis while addressing both privacy and competitive concerns. Despite its clear benefits for the scientific community and societal good, this area remains surprisingly under-explored in the existing literature. From an ML perspective, the current standard training paradigm for data-driven retrosynthesis models is to use one global model trained on a large dataset, \ie, learning with centralized data~\cite{coley2017computer,baylon2019enhancing,yan2020retroxpert,sun2021towards,wan2022retroformer,fang2023single}. Nonetheless, this straightforward training regime has two major flaws. Firstly, its performance is limited by the availability, amount, and diversity of annotated reaction data which might be sensitive or proprietary. Secondly, one single global model may not capture the specific characteristics or preferences of each chemical entity, leading to suboptimal prediction performance~\cite{zhang2021personalized,huang2021personalized}. 

To address these challenges, we propose CKIF (chemical knowledge-informed framework), a privacy-preserving learning framework that enables collaborative learning among chemical entities without sharing raw reaction data. During the learning process, CKIF maintains data anonymity and only the model parameters are communicated. In terms of data heterogeneity (\eg, different data distributions caused by research focuses and industrial interests), CKIF further investigates the distributed learning paradigm from a client-centric perspective~\cite{wang2019federated,huang2021personalized} and thus learns a personalized model for each client. This necessitates the assessment of a model's effectiveness for different clients, as such evaluations can provide dependable guidance to obtain personalized models. In response, CKIF is equipped with a chemical knowledge-informed weighting (CKIW) strategy that harnesses symbolic knowledge representations —-- specifically, molecule fingerprints like extended-connectivity fingerprint (ECFP)~\cite{rogers2010extended} and molecular access system (MACCS) keys~\cite{durant2002reoptimization}. This strategy adaptively adjusts weights during model aggregation to better align the model with the specific preferences of each chemical entity. Experiments on a variety of reaction datasets demonstrate that CKIF outperforms locally trained models, and even models trained on centralized data. In a nutshell, CKIF ensures data privacy and enhances efficiency by distributing computations across multiple clients. It also supports the training of personalized models for each client and leverages reaction data from diverse clients for scalable model training. These benefits make CKIF a feasible and cost-effective approach to advancing retrosynthesis research in a privacy-aware setting.

To summarize, this study underscores the critical need for privacy-aware frameworks in retrosynthesis learning by presenting the first systematic investigation of privacy-preserving methodologies in this domain. We introduce a holistic evaluation framework encompassing curated dataset assembly, rigorous experimental protocols, and benchmarking of mainstream algorithms. To foster transparency and community-driven advancements, the entire codebase is open-sourced. Methodologically, we devise a chemical knowledge-informed weighting strategy that dynamically assigns adaptive weights to evaluate the relevance of external client models, enhancing collaboration efficiency. By fusing domain-specific insights with federated learning principles, this work strengthens the technical groundwork for decentralized retrosynthesis research and promotes responsible data handling. Amidst a global climate of increasing scrutiny over international data sharing, exemplified by recent security directives in fields like healthcare, CKIF's privacy-preserving approach is particularly timely.

\section{Results}\label{sec:res} 

\noindent\textbf{Privacy-aware retrosynthesis with CKIF.}
We prioritize the privacy of chemical reaction data by ensuring that sensitive information is not shared among participants. Instead of gathering raw reaction data into one single edge, CKIF supports distributed model training through the sharing of processed, non-sensitive chemical knowledge (\ie, model parameters) derived from reaction data. CKIF operates through iterative communication rounds, each comprising two stages: local learning and chemical knowledge-informed model aggregation. During local learning, clients independently train their models on proprietary reaction data, initializing from either random value in the first round or from their personalized aggregated model from the previous communication round. In the aggregation stage, clients exchange trained model parameters to leverage collective knowledge while maintaining data privacy. The core of CKIF is its CKIW strategy, which replaces conventional fixed-weight averaging methods. When a client receives models from other participants, it evaluates their relevance using local proxy reaction data and molecular fingerprint similarity metrics. Specifically, each client computes similarities between reactants predicted by other clients' models and ground truth reactants using molecular fingerprints, generating adaptive weights that quantify the value of other clients' models for its specific learning objectives. These chemistry-aware weights guide model aggregation, creating personalized models that reflect each client's preference while benefiting from the collective knowledge of all participants.

\begin{figure*}[t]
    \centering
    \includegraphics[width=\textwidth]{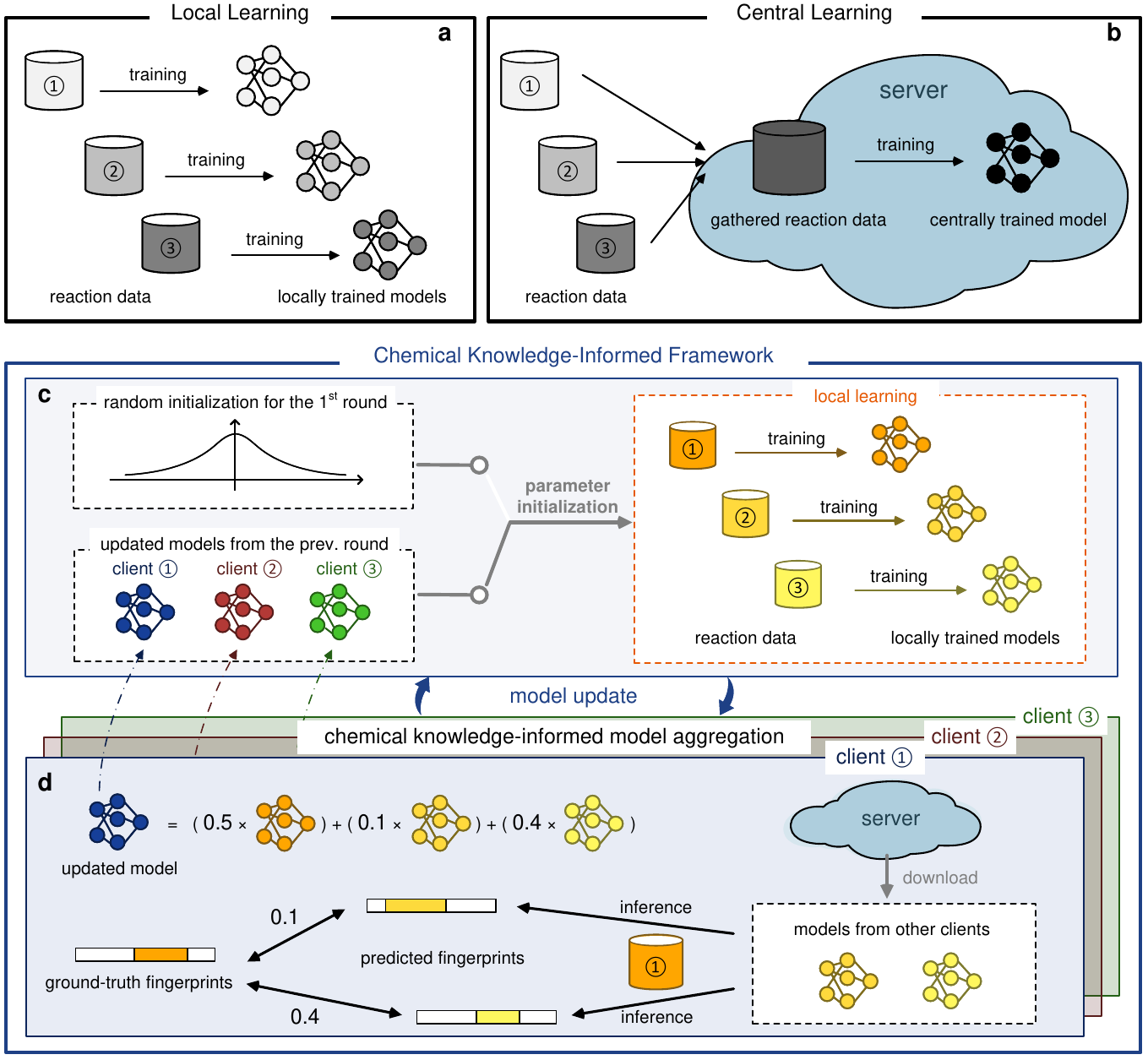}
    \caption{\textbf{Overview of the existing training paradigms and the proposed CKIF (chemical knowledge-informed framework)}. \textbf{a}, Local learning involves individual clients training a model independently on their own reaction data, without interaction or data exchange with other clients.  \textbf{b}, Central learning trains a global model with access to reaction data from all clients. It is the current standard learning paradigm for retrosynthesis. Our privacy-preserving learning process operates through iterative communication rounds, each comprising two stages: local learning (\textbf{c}) and chemical knowledge-informed model aggregation (\textbf{d}). \textbf{c}, In the first round, the model parameters are initialized randomly, while in subsequent communication rounds, they are initialized from the personalized aggregated model obtained in the previous round. \textbf{d}, CKIF incorporates chemical knowledge, \ie, molecule fingerprints, to calculate adaptive weights based on the similarity of molecular fingerprints. These weights are used to guide the model aggregation process.}
    \label{fig:overview}
\end{figure*}

\noindent\textbf{Improvements with CKIF.}
We evaluate the performance of CKIF on a variety of reaction datasets. Top-\textit{K} accuracy, which indicates the proportion of ground truth reactants that appear in (exactly matched) the top-\textit{K} confident predicted reactants, is used for evaluation. Two additional metrics also reported to comprehensively evaluate models' performance. The first is Maximal fragment (MaxFrag) accuracy~\cite{tetko2020state,zhong2022root}, a relaxed version of {top-\textit{K}} accuracy, which focuses on main compound transformations. The second is RoundTrip accuracy~\cite{chen2021deep,wan2022retroformer}, which evaluates whether the predicted reactants can indeed synthesize the target product by utilizing a pre-trained forward synthesis model as an oracle. Tables \ref{tab:t1}, \ref{tab:t2}, and \ref{tab:t3} report the aforementioned evaluation metrics of different methods, namely, Locally Trained (models trained on their single local dataset, see Fig.~\ref{fig:overview}a), Centrally Trained (a single global model trained on all reaction datasets combined, see Fig.~\ref{fig:overview}b), FedAvg~\cite{mcmahan2017communication} (a strong, commonly-used federated baseline), and our proposed method, CKIF (which explicitly models the personalized target distribution for each client, and uses chemical knowledge-informed model aggregation to learn personalized models, see Fig.~\ref{fig:overview}cd). Retrosynthesis learning, as a seq2seq problem, poses unique challenges that limit the direct applicability of many state-of-the-art (SOTA) federated learning methods, which are often designed for classification or regression tasks~\cite{tan2022towards}. We selected FedAvg as our primary baseline due to its proven efficacy in seq2seq contexts~\cite{lin2022fednlp}, and its widespread adoption in both academic and industrial settings. In addition, adapting other SOTA methods to seq2seq tasks would require significant re-engineering, potentially introducing unfair comparisons. Note that Centrally Trained with full access to all reaction data may bring significant privacy concerns.

The experimental results in Table~\ref{tab:t1} evidence that CKIF consistently outperforms Locally Trained across four clients sampled from USPTO-50K dataset~\cite{schneider2016s}. Taking client $C_2$ as an example, Locally Trained performs poorly with an accuracy of 4.1\% at \textit{K}=1, whereas CKIF shows a significant improvement, achieving an accuracy of 23.6\%. This substantial enhancement suggests that the proposed CKIF effectively leverages the chemical knowledge-guided model aggregation to enhance the performance over the Locally Trained baseline. In addition, CKIF exhibits comparable or even superior performance compared to the centrally trained model in several cases. For example, CKIF outperforms Centrally Trained on client $C_3$ by up to 8.8\%. This highlights the potential of CKIF in leveraging the personalized target distribution and CKIW-based model aggregation  compared to training a single global model on all reaction datasets combined. FedAvg, a strong baseline used in federated learning, treats all clients equally and aggregates updates from them to create a global model, ignoring personalized target distributions of each client. In contrast, CKIF explicitly models the non-independent and identically distributed nature for each client. As a result, CKIF demonstrates considerable performance advantages over FedAvg.

\noindent\textbf{Scalability to number of clients.}
To investigate the effect of increasing the number of clients, we conduct experiments using eight clients sampled from USPTO-50K~\cite{schneider2016s} and USPTO-MIT~\cite{jin2017predicting}. The experimental results in Table~\ref{tab:t2} demonstrate that as the number of clients increases, the performance of all clients further improves, highlighting the scalability of CKIF. For example, CKIF achieves a top-1 accuracy of 30.9\% with four participants and an improved one of 36.9\% with eight participants in terms of $C_3$. This finding suggests that the collaborative nature of CKIF enables the extraction of beneficial guidance from a larger pool of clients, leading to enhanced performance for all participating clients.

In addition, the experimental results reveal a counterintuitive decline in performance for Centrally Trained as data volume increases, challenging the common assumption that more data invariably improves performance~\cite{edunov2018understanding,sugiyama2019data}. We attribute this to three factors: lack of specialization, distribution mismatch, and bias towards majority reaction data. The centrally trained model, while capturing overall trends, may overlook client-specific characteristics. It also faces potential distribution mismatches between diverse training reaction data and distributions of individual chemical entities. In addition, reaction data from certain clients may dominate the training process, potentially biasing the model. In contrast, our CKIF maintains individual models for each client, explicitly modeling both global consistency and personalized preferences. Such a framework enables better alignment with client-specific data distributions, yielding improved performance.

\begin{figure*}[b]
    \centering
    \includegraphics[width=\textwidth]{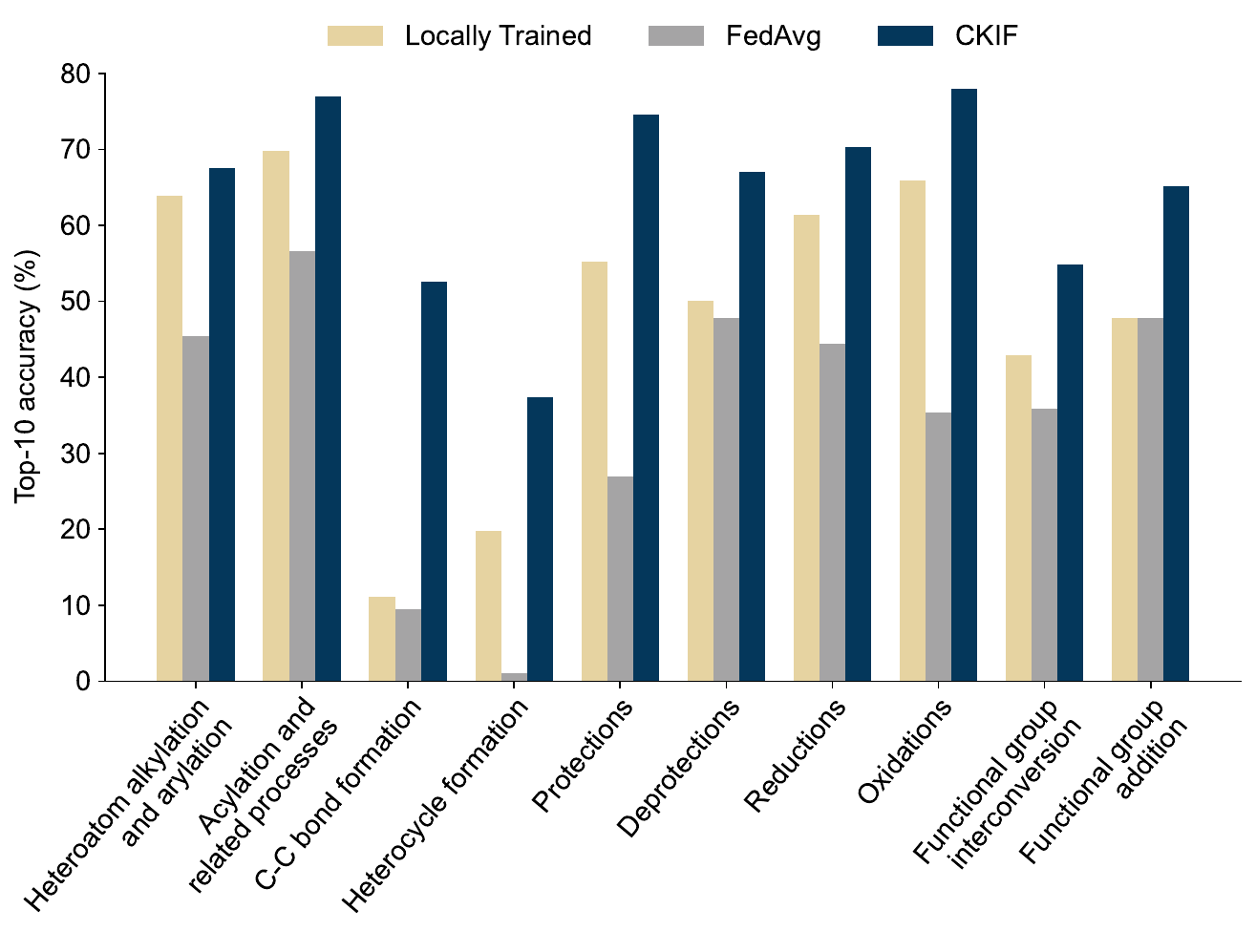}
    \caption{\textbf{Comparative analysis of prediction performance for different reaction classes}. The figure shows the Top-10 accuracy of three methods (Locally Trained, FedAvg~\cite{mcmahan2017communication}, and CKIF) across 10 major classes of chemical transformations in the USPTO-50K dataset. The results demonstrate that CKIF (chemical knowledge-informed framework) outperforms the strong baseline by a considerable margin on all reaction types. Source data are provided as a Source Data file.}
    \label{fig:rea_type}
\end{figure*}

\noindent\textbf{Analysis of accuracy for different reaction types.} 
In the realm of chemistry, reaction classes serve as invaluable tools for chemists to navigate vast databases of reactions and extract similar members within the same class for analysis and determination of optimal reaction conditions. Moreover, reaction classes offer a concise and efficient means of communication, enabling chemists to describe the functionality and underlying mechanisms of chemical reactions in terms of atomic rearrangements. USPTO-50K dataset provides high-quality annotations which assigns one of ten reaction classes to each reaction, facilitating further investigation into the performance gains associated with specific reaction classes. From the results shown in Fig. \ref{fig:rea_type}, we can observe that CKIF outperforms the Locally Trained and FedAvg baselines by a considerable margin among all reaction classes and achieves excellent results. Also, we find that C-C bond formation is the easiest to improve among the ten reaction classes, while heteroatom alkylation and arylation is the most challenging one. The reasons are two folds: 1) C-C bond formation has more diverse possibilities for choosing reactants and reactions than other reaction classes~\cite{tetko2020state}, while the reaction data utilized by Locally Trained is insufficient to learn such transformations. 2) There are sufficient reaction data of heteroatom alkylation and arylation for Locally Trained to learn. Note that reaction classes are usually unknown in the real world, we do not utilize such information in all experiments as~\cite{zhong2022root}.

\noindent\textbf{Scalability to number of training samples.} 
Tables \ref{tab:t1} and \ref{tab:t2} demonstrate the effectiveness and scalability of the CKIF compared to several strong baselines. We can observe that some clients suffer from limited data volume, \eg, top-1 accuracy of 4.1\% on $C_2$. Such data scarcity problem can be mitigated with more reaction data in the same distribution. Although collecting data is expensive and time-consuming, one may wonder whether doing so can lead to better prediction performance. To gain insights into the performance gains of CKIF when clients have sufficient data volume and further validate scalability of CKIF, we conduct experiments on USPTO 1k TPL dataset~\cite{schwaller2021mapping}, where the clients have larger data volume. 

From the experimental results in Table \ref{tab:t3}, we have the following findings: 1) With a sufficient volume of client data, all the methods, including Locally Trained, Centrally Trained, FedAvg, and CKIF, show impressive performance compared to previous experiments. This indicates that the availability of a larger volume of data allows models to learn sufficient underlying patterns so as to enhance their retrosynthesis prediction accuracy. 2) Despite considerable performance of Locally Trained models with sufficient client data, the experimental results show that CKIF can still further improve retrosynthesis accuracy. For instance, CKIF achieves an average performance improvement of $+$6.9\% and $+$16.6\% on top-1 accuracy compared to locally and centrally trained models, respectively. In a nutshell, the experimental results with larger data volume show that CKIF can further improve the performance even when clients have sufficient data volume to train their own personalized models. CKIF consistently outperforms the Locally Trained baseline and brings performance benefits to clients with both low and high data volumes. This reaffirms the scalability and effectiveness of CKIF in leveraging chemical knowledge and personalized target distribution for enhanced retrosynthesis accuracy.

\begin{figure*}[t]
    \centering
    \includegraphics[width=\textwidth]{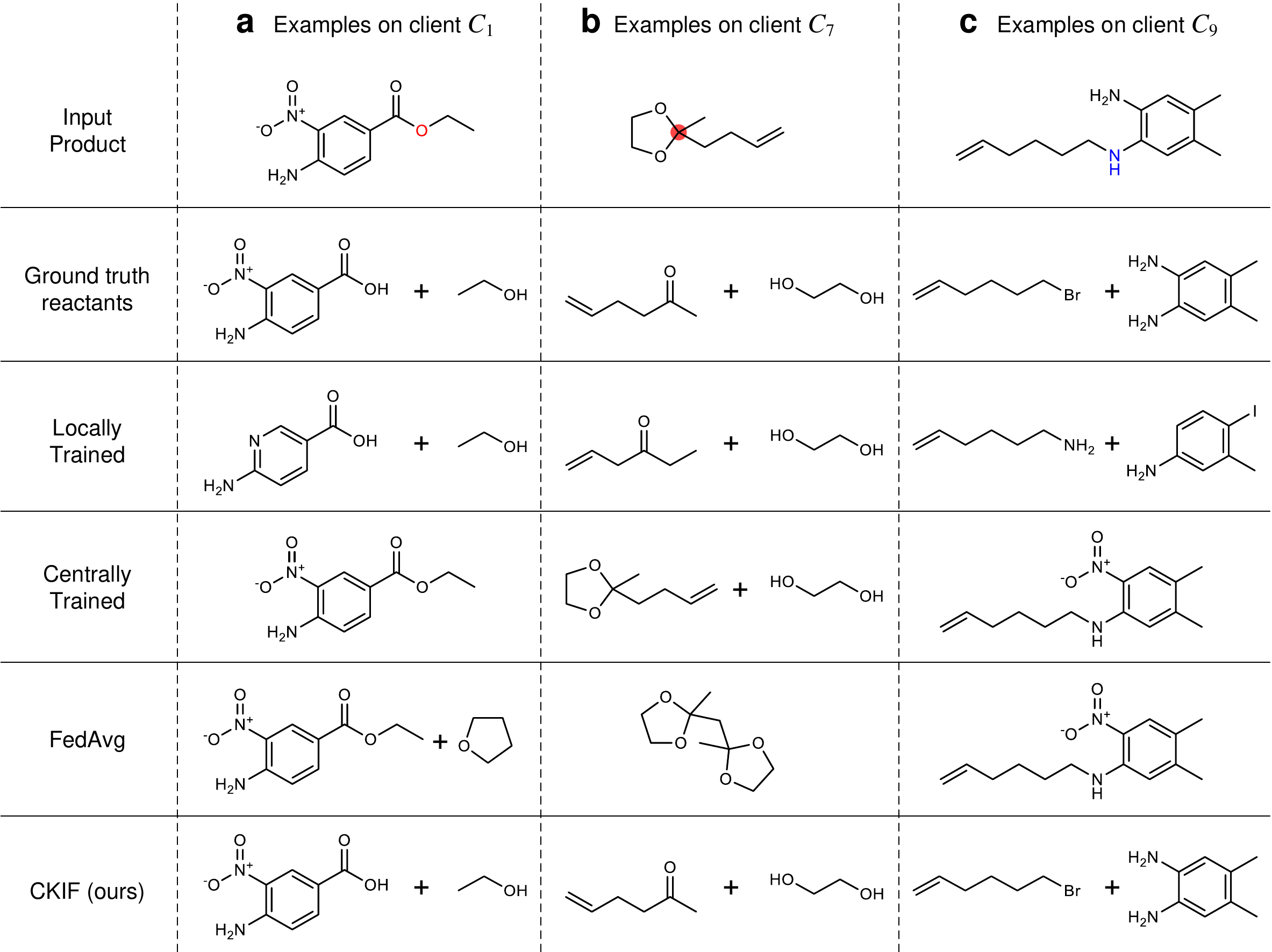}
    \caption{\textbf{Qualitative comparison of retrosynthesis predictions by CKIF (chemical knowledge-informed framework), FedAvg~\cite{mcmahan2017communication}, Centrally Trained, and Locally Trained.} 
    Representative examples are shown for retrosynthesis predictions on three distinct input products (\textbf{a}-\textbf{c}). For each case, the target product is displayed at the top, followed by the ground truth reactants and the predictions from four different methods. The examples are drawn from diverse datasets corresponding to different clients: \textbf{a} an acylation-related reaction from the USPTO-50K dataset (client $C_1$), \textbf{b} a ring formation reaction from the USPTO-MIT dataset (client $C_7$), and \textbf{c} a reaction from a mixed subset of the USPTO 1k TPL dataset (client $C_9$). These results qualitatively demonstrate CKIF's superior accuracy and its ability to generalize across complex and varied reaction types. Different reactive sites are highlighted with different colors.}
    \label{fig:qualitative}
\end{figure*}

\noindent\textbf{Qualitative analysis.}  
To compare the qualitative results of these strong baselines and CKIF, we showcase three reactions with their characteristics (reaction types), as depicted in Fig. \ref{fig:qualitative}. Experimental results indicate that CKIF consistently outperforms the other baselines and provides plausible synthesis routes. Taking the ring formation reaction (Fig. \ref{fig:qualitative}b) as an example, the ability of CKIF to comprehend the structural requirements for forming rings leads to correct predictions, while the other baselines struggle to predict the ring formation, resulting in incorrect reactants. The superior performance of CKIF highlights the efficacy of leveraging both local (personalized model for each client) and global (client-centric weighting strategy) information to achieve accurate retrosynthetic predictions in a privacy-aware setting.

\begin{figure*}[t]
    \centering
    \includegraphics[width=\textwidth]{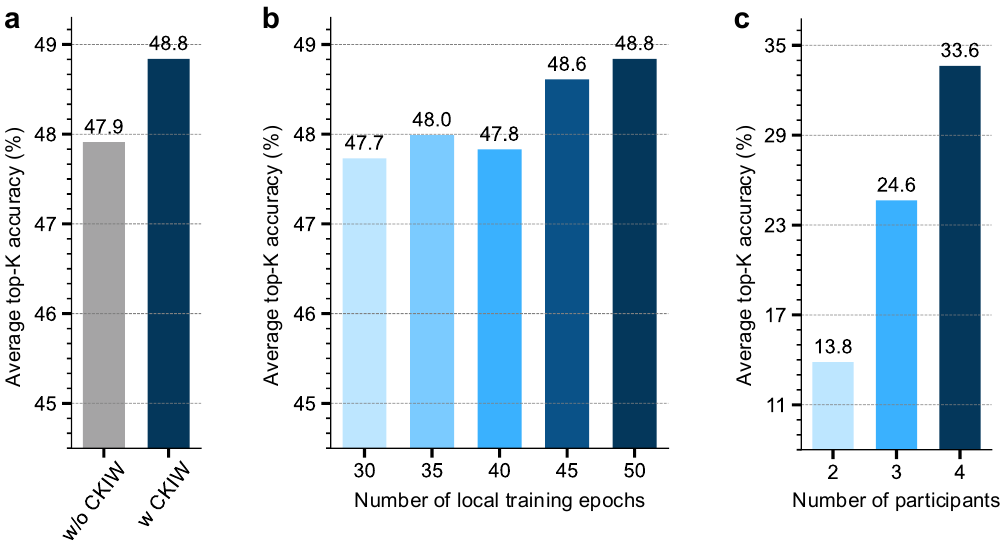}
    \caption{\textbf{Ablative experiments on the USPTO-50K dataset}, including \textbf{a} performance comparison between CKIW (chemical knowledge-informed weighting) and w/o CKIW (count-based averaging baseline), \textbf{b} trends in average top-\textit{K} accuracy with increasing local training epochs, and \textbf{c} trends in average top-\textit{K} accuracy with increasing clients participating in the federated learning process. Source data are provided as a Source Data file.}
    \label{fig:ab_exp}
\end{figure*}

\noindent\textbf{Analysis of the chemical knowledge-informed weighting strategy.}  
To ascertain the effectiveness of our proposed CKIW strategy, we conducted a comparative analysis against a robust baseline: Average Aggregation, which uses average model aggregation instead of using the proposed chemical knowledge-informed model aggregation. The results shown in Fig.~\ref{fig:ab_exp}a demonstrate a significant enhancement; CKIF improved predictive accuracy by $+$0.93\% in terms of average top-\textit{K} accuracy. Given that this improvement is averaged over 16 metrics for 4 clients in USPTO-50K, it illustrates the considerable benefit of CKIF. This analysis highlights the substantial benefits of incorporating chemical knowledge into model aggregation, suggesting a powerful alternative to heuristic methods such as count-based model averaging.

\noindent\textbf{Sensitivity analysis.}
After verifying the effectiveness of CKIW, we conduct ablative experiments to assess the robustness of CKIF under varying learning settings. In particular, we explore the impact of two critical hyperparameters: the number of communication rounds and the number of clients. The resilience of CKIF to the number of communication rounds is tested with values set at 30, 35, 40, 45, and 50. The results, as illustrated in Fig.~\ref{fig:ab_exp}b, show a gradual improvement in model accuracy from 47.73\% to 48.84\% as the number of epochs increased. Similarly, we investigate the effect of varying the number of clients participating in the learning process. The configurations tested include 2, 3, and 4 clients, with corresponding accuracies of 13.85\%, 24.65\%, and 33.63\% (Fig.~\ref{fig:ab_exp}c). This progression demonstrates a robust scaling behavior of CKIF as the network grows, highlighting its capability to harness chemical knowledge across more clients. These findings underscore the resilience and scalability of CKIF, affirming its effectiveness across diverse experimental settings.

\begin{figure*}[t]
    \centering
    \includegraphics[width=\textwidth]{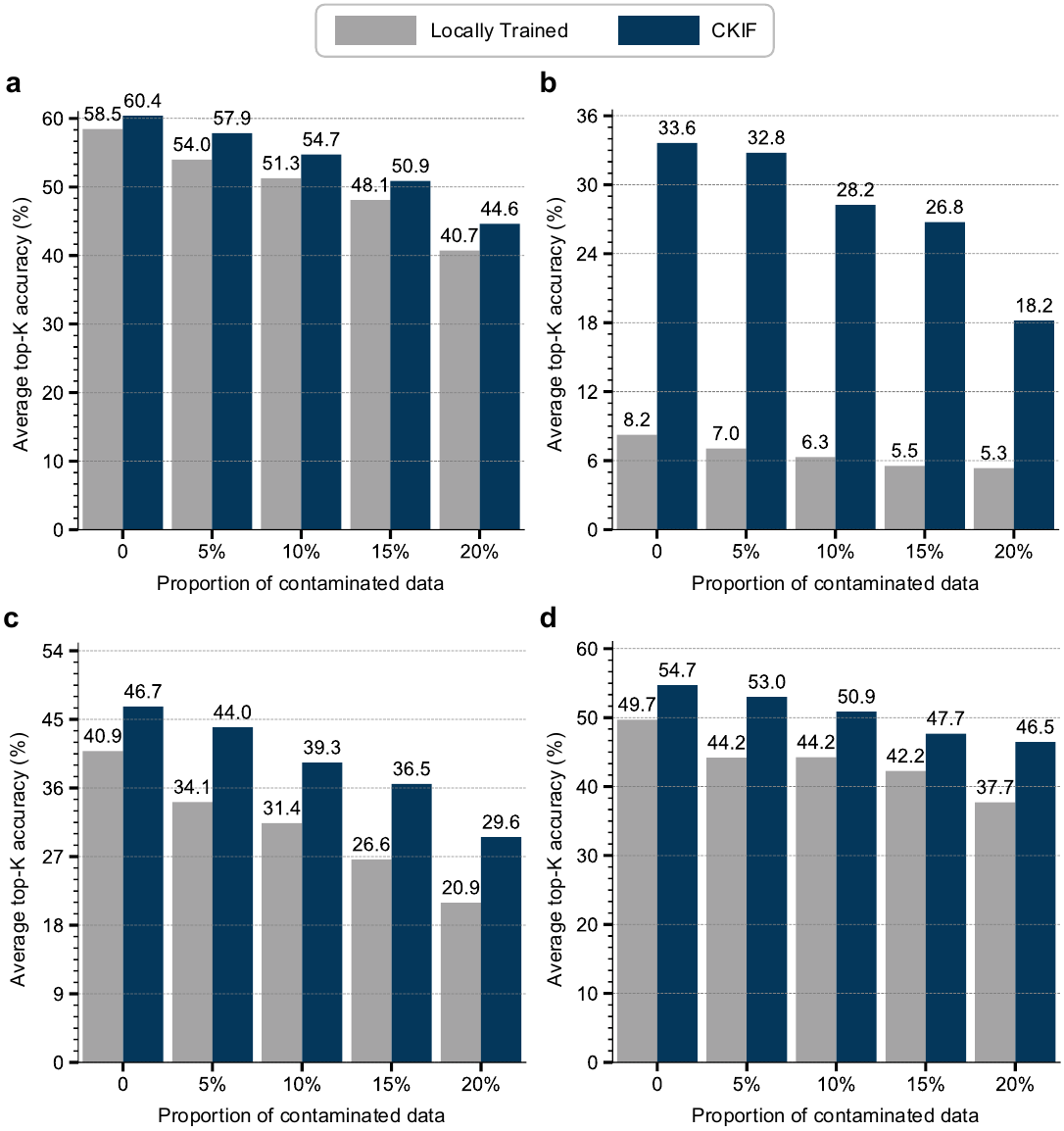}
    \caption{\textbf{Robustness of CKIF (chemical knowledge-informed framework) to data contamination across heterogeneous clients}. The figure shows a robustness analysis comparing the CKIF framework (dark blue) with the Locally Trained baseline (grey). The four panels (\textbf{a}-\textbf{d}) show the performance of four participating clients ($C_1$-$C_4$), respectively. These results demonstrate that CKIF consistently outperforms Locally Trained for all clients and at every level of data contamination. Source data are provided as a Source Data file.}
    \label{fig:ab_robust}
\end{figure*}

\noindent\textbf{Robustness analysis.}
The robustness of CKIF against contaminated client data is evaluated by investigating the scenario with contamination across all participating clients. Following~\citet{toniato2021unassisted}, we simulate contamination (\ie, chemically wrong entries) by randomly exchanging reactants and products while shuffling SMILES tokens. To ensure fair comparison, the test split of all clients remains unchanged.  Results (Fig.~\ref{fig:ab_robust}) demonstrate that CKIF maintains a significant performance advantage over Locally Trained models across varying contamination levels. Even with 5$\sim$10\% contaminated data, CKIF might achieve comparable performance to that of Locally Trained models trained on clean data. The core reason is that CKIW dynamically adjusts each client's contribution during model aggregation so as to mitigate the impact of contaminated data. These findings highlight CKIF's inherent robustness to data contamination. Note that CKIF is not immune to malicious attacks such as data poisoning. While the comprehensive handling of contaminated client data and defense mechanisms against malicious attacks warrant thorough investigation, they extend beyond the scope of this work and are addressed in the Discussion section.

\section{Discussion}\label{sec:dis}

As an essential skill for organic chemists, retrosynthesis requires expert knowledge, creativity, and intuition. In the modern era, chemical reaction data become a valuable asset across various scientific domains such as materials science, pharmaceuticals, and energy. However, sharing such data can expose trade secrets, violate patents, and lead to unfair competition or security compromises. To address these challenges, this paper introduces CKIF, a distributed machine learning framework that enables collaborative training without the need to exchange raw reaction data, thereby ensuring data privacy --- a major concern in sensitive scientific domains. By leveraging the distributed and heterogeneous nature of reaction data sources, CKIF not only improves efficiency but also reduces communication overhead, latency, and bandwidth consumption. CKIF allows for the distribution of computation across multiple clients closer to data sources, enhancing parallelism and fault tolerance.

CKIF incorporates chemical symbolic knowledge in the form of molecule fingerprints to guide model aggregation, resulting in personalized models that enhance the accuracy and relevance of retrosynthesis predictions. This knowledge-based approach allows each participating chemical entity to benefit from the collective insights. In addition, empirical results suggest that CKIF's performance improves with an increasing number of clients and training samples, showcasing CKIF's scalability. By applying CKIF to retrosynthesis, one can expect the collaborative discovery and optimization of new chemical transformations from various and confidential data sources. Experiments on various experimental settings show that CKIF outperforms locally trained models (Locally Trained) and even those trained on centralized data (Centrally Trained). We hope that CKIF opens up new possibilities for advancing privacy-aware retrosynthesis and related fields. In addition, we aim to inspire more research on privacy-preserving ML methods for other scientific domains and applications.

Despite these strengths, our work has limitations that necessitate further exploration. First, CKIF'performance relies on the quality of local data from each chemical entity, which might vary significantly in practice. Second, the model aggregation based on chemical knowledge uses the same molecule fingerprints and measurements for all clients, which might not accurately capture the personalized target distribution of each client and might contain unintentional bias. Future work will explore customized molecule fingerprints or measurements that meet the unique needs of each client. Third, our evaluation metrics are focused on the validity and uniqueness of the generated synthetic routes, overlooking the practicality and feasibility of the synthesis in terms of cost, yield, and environmental impact.

During the privacy-aware learning process, each chemical entity trains their personalized models locally and only model parameters are communicated. However, this process is not immune to privacy threats, as malicious attackers can infer sensitive information from the model updates exchanged between the clients~\citep{xie2019dba,lyu2022privacy}, \ie, data representation leakage from gradients or models. Several methods have been proposed to address this essential issue of privacy leakage, such as encrypting the model updates or adding noise to them using differential privacy~\citep{abadi2016deep,sun2021provable}. While addressing privacy leakage and other security concerns (such as data poisoning) remains an active research area~\citep{tolpegin2020data,sun2021data,chen2024federated,sharma2024review,nowroozi2025federated}, our contribution is orthogonal to those defense developments. Defense methods, in principle, can be seamlessly incorporated into our framework.

\section{Methods}\label{sec:met}

\textbf{Problem formulation.}
In the present study, we focus in particular on the challenge of learning retrosynthesis models in a privacy-aware setting where reaction data are proprietary and cannot be shared among chemical entities (clients) or with an external central server. To address this, CKIF is specifically designed to allow for the collaborative improvement of retrosynthesis models without sharing sensitive reaction data between participants. The collaboration occurs through multiple rounds of communication where only implicit and explicit chemical knowledge --- rather than raw reaction data --- is exchanged. Note that the term ``privacy-aware'' specifically refers to the prevention of raw data sharing --- a fundamental requirement in collaborative retrosynthesis learning where reaction data cannot be directly exchanged between entities. This aligns with the minimal privacy definition widely adopted in applied FL studies~\cite{ogier2023federated,karargyris2023federated,feng2024robustly}. For the retrosynthesis prediction task itself, we define it as the process of training a machine learning model (or hypothesis) that maps a target molecule to its corresponding precursors (\ie, reactants).

\noindent\textbf{Privacy-aware model training.} 
Transformer~\cite{vaswani2017attention} is adopted to learn a probabilistic mapping $P(\mathbf{y} \vert \mathbf{x})$ from a given target molecule $\mathbf{x}$ to reactants $\mathbf{y}$. Suppose there are a total of $K$ clients, each possessing a local reaction dataset $\{(\mathbf{x}_n, \mathbf{y}_n)\}_{n=1}^{N_i}$, where $N_i$ denotes the number of reactions stored on the $i$-th client $C_i$. Our objective is to collaboratively learn a retrosynthesis model with parameters $\boldsymbol{\uptheta}$, without centralizing or sharing training reaction data. In the remaining content, we first formally define the steps of each communication round of CKIF, and introduce the learning objective of retrosynthesis model. We then elaborate on the CKIW strategy which learns a personalized model for each client.

At each communication round of CKIF, three procedures are performed sequentially for all clients. Firstly, model parameters generated by the previous round are used for parameter initialization (random initialization for the first round). Secondly, the initialized models of each client are trained locally on their own reaction data for a fixed number of epochs (or iterations) to minimize the following objective function:
\begin{equation}
    \mathcal{L} = -\log P(\mathbf{y} \vert \mathbf{x};\boldsymbol{\uptheta}).
\end{equation}
Once trained, each client $C_i$ sends their parameters $\boldsymbol{\uptheta}_{i}$ to the central server. Thirdly, after collection, these parameters are aggregated (\ie, weighted sum) to a new global model by the central server: $\boldsymbol{\uptheta}_{\mathrm{G}} = \sum_{k=1}^{K} w_k \boldsymbol{\uptheta}_{k}$, where $w_k$ is the weighting factor for client $C_k$. As one of the most widely used methods in federated learning, FedAvg~\citep{mcmahan2017communication} defines the weighting factor for $C_i$ as the proportion of training samples: $w_i = N_i / \big(\sum_{k=1}^{K}N_k\big)$.

Despite being prevalent, mainstream federated learning methods like FedAvg struggle to handle data heterogeneity~\cite{wang2019federated,huang2021personalized} --- one common characteristic of chemical entities, where reaction data often reflects specialized research focuses or industrial interests. To address this, CKIF advances privacy-preserving retrosynthesis learning by prioritizing client-specific requirements. Our method equips each client with a personalized model, tuned to their local data distribution and objectives, while still leveraging insights from other participants through the privacy-preserving exchange of model parameters alone. Critically, CKIF replaces the standard aggregation process, which yields a single global model ($\boldsymbol{\uptheta}_{\mathrm{G}}$), with a tailored aggregation mechanism that constructs a unique, personalized model for each client in every communication round (see Box 1). For instance, the personalized model for client $C_i$ is obtained by:
\begin{equation}
    \boldsymbol{\uptheta}_{i} = \sum\nolimits_{k=1}^{K} w_{i,k}^a \boldsymbol{\uptheta}_{k},
\end{equation}
where $w_{i,k}^a$ denotes the adaptive weighting factor power by the proposed CKIW strategy (Box 2). $w_{i,k}^a$ quantifies the importance of models from other clients $C_k$ relative to the given client $C_i$, and is normalized as:
\begin{equation}
    w_{i,k}^a=\begin{cases}\mu,& i = k \\
    (1-\mu)\frac{\exp(s_{i,k} / \tau)}{\sum_{j=1,i\neq j}^{K}\exp(s_{i,j}/ \tau)},& i \neq k\end{cases},
\end{equation}
where $\mu$ and $\tau$ are hyperparameters that control the influence of self's model $\boldsymbol{\uptheta}_{i}$ and neighboring clients' models $\boldsymbol{\uptheta}_{k}$. $s_{i,k}$ represents the similarity score between the reactants predicted by $\boldsymbol{\uptheta}_{k}$ and the annotated ones stored in $C_i$. Note that the process of measuring similarity involves using proxy reaction data, which are stored securely and processed locally on each client. During the collaborative training process, only the model parameters are communicated. In our experiment, all reaction data in the validation set are chosen as the proxy reaction data. Clients can optionally provide additional proxy reaction data for improved measurement.

\clearpage
\begin{mybox}{\small Box 1 \textbar ~Algorithm of chemical knowledge-informed framework ~~~~~~~~~~~~~~~~~$_{\!\!}$~}
    \vspace{3.5mm}
    \small
    \begin{algorithmic}[1]
        \Require{Number of chemical entities $K$, number of communication rounds $R_T$, local epochs $R_L$, number of fine-tuning rounds $R_F$}
        \Ensure{Personalized models $\{\boldsymbol{\uptheta}_{1}, \boldsymbol{\uptheta}_{2}, ..., \boldsymbol{\uptheta}_{K}\}$}
        \State Initialize models $\{\boldsymbol{\uptheta}_{1}^0, \boldsymbol{\uptheta}_{2}^0, ..., \boldsymbol{\uptheta}_{K}^0\}$
        \For{$t = 1$ to $R_T$}
            \For{$k = 1$ to $K$} 
                \State $\boldsymbol{\uptheta}_{k}^t \leftarrow \text{LocalLearning}(\boldsymbol{\uptheta}_{k}^{t-1}, R_L)$
            \EndFor
            \State Compute $\{w_{i,k}^a\}$ with chemical knowledge guidance (Box 2)
            \For{$i = 1$ to $K$}
                \State $\boldsymbol{\uptheta}_{i}^t \leftarrow \sum\nolimits_{k=1}^{K} w_{i,k}^a \boldsymbol{\uptheta}_{k}^t$
            \EndFor
        \EndFor
        \For{$k = 1$ to $K$} 
            \State $\boldsymbol{\uptheta}_{k} \leftarrow \text{LocalLearning}(\boldsymbol{\uptheta}_{k}^{R_T}, R_L \cdot R_F)$
        \EndFor
        \State \Return $\{\boldsymbol{\uptheta}_{1}, \boldsymbol{\uptheta}_{2}, ..., \boldsymbol{\uptheta}_{K}\}$
        
        \end{algorithmic}
\end{mybox}

\begin{mybox}{\small Box 2 \textbar ~Algorithm of  chemical knowledge-informed weighting strategy ~~~~~~~$_{\!\!\!}$~}
    \vspace{3mm} 
    \small 
    \begin{algorithmic}[1]
        \Require{Client models $\{\boldsymbol{\uptheta}_{1}, \boldsymbol{\uptheta}_{2}, ..., \boldsymbol{\uptheta}_{K}\}$, proxy reaction data $\{D_1, D_2, ..., D_K\}$, hyperparameters $\mu$, $\tau$}
        \Ensure{Adaptive weighting factors $\{w_{i,k}^a\}$}
        \For{$i = 1$ to $K$}
            \For{$k = 1$ to $K$}
            \Comment{Client $C_i$ evaluates model $\boldsymbol{\uptheta}_{k}$ locally on its own proxy data without sharing}
                \If{$i = k$}
                    \State $w_{i,k}^a \leftarrow \mu$
                \Else
                    \State $s_{i,k} \leftarrow 0$
                    \For{$(\mathbf{x}, \mathbf{y}) \in D_i$}
                        \State $\hat{\mathbf{y}} \leftarrow \text{PredictReactants}(\mathbf{x}, \boldsymbol{\uptheta}_{k})$
                        \State $f_{\mathbf{y}} \leftarrow \text{ExtractFingerprint}(\mathbf{y})$
                        \State $f_{\hat{\mathbf{y}}} \leftarrow \text{ExtractFingerprint}(\hat{\mathbf{y}})$
                        \State $s_{i,k} \leftarrow s_{i,k} + \text{MolecularSimilarity}(f_{\mathbf{y}}, f_{\hat{\mathbf{y}}})$
                    \EndFor
                    \State $s_{i,k} \leftarrow s_{i,k} / |D_i|$
                \EndIf
            \EndFor
            \For{$k = 1$ to $K$, $k \neq i$}
                \State $w_{i,k}^a \leftarrow (1-\mu)\frac{\exp(s_{i,k} / \tau)}{\sum_{j=1,j\neq i}^{K}\exp(s_{i,j}/ \tau)}$
            \EndFor
        \EndFor
        \State \Return $\{w_{i,k}^a\}$
    \end{algorithmic}
\end{mybox}

\clearpage

In the above distributed training procedure, the strategy used to assign weights to individual models during aggregation is of great importance~\cite{zhang2021personalized} for the performance, convergence, \etc. Generic weighting strategies, such as count-based averaging, rely on handcrafted metrics. Such heuristic methods have significant limitations: they ignore the potential heterogeneity in data distributions across different chemical entities. This raises a fundamental question: how to compute meaningful weights $s_{i,k}$ that quantify how valuable one client's model is for another client's learning objectives? To address this, CKIF is equipped with the CKIW strategy that harnesses the explicit, symbolic chemical knowledge --- specifically, molecule fingerprints like ECFP~\cite{rogers2010extended} and MACCS keys~\cite{durant2002reoptimization}. This is inspired by~\citet{asahara2022extended}, who demonstrated fingerprints' utility as chemical reaction representations for reaction prediction tasks. By calculating the molecule similarity between predicted and actual reactants, CKIW assesses the model's effectiveness (\ie, $s_{i,k}$) across reaction data of different clients. The computation procedure of $s_{i,k}$ is detailed as follows. For each reaction pair $(\mathbf{x},\mathbf{y})$ stored on the validation set of $C_i$, we first get the predicted reactants $\hat{\mathbf{y}}$ through the personalized model $\boldsymbol{\uptheta}_{k}$. Then, molecule fingerprints~\cite{rogers2010extended,durant2002reoptimization} of $\mathbf{y}$ and $\hat{\mathbf{y}}$ are extracted by RDKit~\cite{landrum2013rdkit}. Thereby, we can measure the similarity of the molecules by analyzing the molecule fingerprints with established measurement methods~\cite{bajusz2015tanimoto}. The selection of measurement methods and molecule fingerprints is flexible, allowing CKIW to take advantage of advancements in these areas. Finally, we can obtain $s_{i,k}$ by averaging molecule similarities over all pairs in the proxy reaction data. In summary, $s_{i,k}$ represents the overall performance of $\boldsymbol{\uptheta}_{k}$ on ${C_i}$'s proxy data, with higher values indicating greater similarity between the clients' data distributions. Note that throughout the aggregation process, each client $C_i$ only uses their own proxy data locally to evaluate other clients' models. The only information shared between clients are the model parameters $\boldsymbol{\uptheta}_{k}$, ensuring that private reaction data remains protected.

\noindent\textbf{Theoretical analysis.} We present the theoretical analysis of the proposed CKIF framework, which covers both the optimization error and the generalization error. The optimization error measures the convergence rate of the CKIF on the seen data during training. The generalization error measures the performance of the personalized model on the unseen data, which is more fundamental in the machine learning community. Although Adam is adopted in the experiments, the theoretical analysis of CKIF is conducted using Local SGD \citep{stich2018local}. This preserves CKIF's fundamental properties while facilitating a more tractable analysis. Considering that the weighting matrix of CKIF is not necessarily symmetric and doubly stochastic, we provide the theoretical analysis in the setting of decentralized stochastic learning with time-varying and row-stochastic weighting matrices. The theoretical results are established in the non-convex and stochastic settings. Compared with previous analyses of decentralized stochastic optimization that assume doubly stochastic mixing matrices \citep{koloskova2020unified}, establishing convergence rates for CKIF is more challenging because its weighting matrices are only row-stochastic and sample-dependent. Moreover, the derived generalization error bounds for the personalized model of CKIF capture the influence of the size of the local dataset and the time-varying weighting matrix, which are comparable to those of its doubly stochastic counterpart in previous work \citep{Hu_Gong_Xu_Liu_Luan_Wang_Liu_2025}.

\noindent\textbf{Implementation details and hyperparameters.} We formulate retrosynthesis as a sequence-to-sequence~\cite{sutskever2014sequence} problem where each element in the sequence is a SMILES token. We use a standardization protocol~\cite{wan2022retroformer,chen2021deep,schwaller2019molecular} using RDKit's canonical SMILES representation (rdkit.Chem.MolToSmiles with canonical=True). This standardization step is applied before fingerprint calculation and model training, ensuring consistent molecular representation across all clients. A standard Transformer~\cite{vaswani2017attention} composed of an encoder-decoder architecture is adopted as the retrosynthesis model. All the baseline models use the same transformer architecture. Specifically, the adopted transformer consists of a 6-layer encoder and a 6-layer decoder, and the number of attention heads is set to 8. We train the Personalized and centrally trained models for 250 epochs with a batch size of 64. The experiments involving 4 and 8 clients were conducted using three RTX 3090 GPUs and four RTX 4090 GPUs, respectively. We employ an Adam optimizer~\cite{kingma2014adam} with a learning rate of 0.0002, $\beta_1= 0.9$, and $\beta_2= 0.998$. Dropout~\cite{srivastava2014dropout} is applied to the whole model to avoid overfitting. For CKIF,  we trained the models for $R_T$ rounds, each with $R_L$ local epochs with a batch size of 64. Inspired by~\cite{wang2019federated,yu2020salvaging}, in the last $R_F$ rounds, we perform local fine-tuning instead of model aggregation. For all experiments, $R_T$, $R_L$, and $R_F$ are set to 50, 5, and 10, respectively. The hyperparameters $\mu$ and $\tau$ are set to be $1 / K $ and 1.5, respectively. MACCS keys~\cite{durant2002reoptimization} and Tanimoto similarity~\cite{bajusz2015tanimoto} are employed to measure the similarity of molecules. OpenNMT-py~\cite{klein2017opennmt}, an open-source neural machine translation framework in PyTorch~\cite{paszke2019pytorch}, is adopted to build our models.

\noindent\textbf{Dataset collection.} 
Our CKIF is evaluated on a variety of federated datasets sampled from U.S. patent database curated by~\cite{lowe2012extraction,Lowe2017} for retrosynthesis prediction and reaction classification: USPTO-50K~\cite{schneider2016s}, USPTO-MIT~\cite{jin2017predicting}, and USPTO 1k TPL~\cite{schwaller2021mapping}.
\begin{itemize}
    \item USPTO-50K is a high-quality reaction dataset comprising approximately $50,000$ reactions, each with manually annotated reaction type. We used the same split introduced in~\cite{liu2017retrosynthetic,dai2019retrosynthesis}, in which 80\% of the data were used for training, 10\% for validation, and 10\% for testing. Five clients are obtained according to the reaction classes. In addition, the reactions with chirality (a property of asymmetry) compose another client.
    \item USPTO-MIT is a larger dataset which contains approximately $480,000$ reactions. $\sim 410,000$ reactions were used for training, $\sim 30,000$ for validation, and $\sim 40,000$ for testing, consistent with the split used in~\cite{jin2017predicting}. Two clients are obtained according to the reaction characteristics, \ie, ring opening and ring formation.
    \item USPTO 1k TPL is a reaction dataset containing the 1,000 most common reaction templates as classes, which is built by~\cite{schwaller2021mapping} for reaction classification. Eight clients are obtained according to the reaction classes (templates).
\end{itemize}
\noindent An overview of the datasets used for training and evaluation, including data split and statistics, is shown in Supplementary Table 1.

\noindent\textbf{Pre- and post-processing.} 
For all chemical reactions from datasets, reactants and products are extracted to form paired reaction data. Following previous works~\cite{segler2017neural,dai2019retrosynthesis,wan2022retroformer}, reagents and edge products are ignored in all experiments. We focus on single-step retrosynthesis~\cite{shi2020graph,yan2020retroxpert,somnath2021learning,wang2021retroprime}, where reactions containing multiple products are split into individual reactions to ensure that each reaction exclusively featured a single product. Canonical SMILES~\cite{weininger1989smiles} is used to represent molecules in all experiments. For post-processing, we use beam search~\cite{freitag2017beam} to systematically explore potential reactant combinations. The ranking of reactant candidates was performed based on the models' confidence scores, which reflected the likelihood of a given reactant combination leading to the desired product.

\noindent\textbf{Evaluation metrics.} 
We use top-\textit{K} ($\textit{K} = \{1,3,5,10\}$) retrosynthesis accuracy, the most widely metric used among current literatures~\cite{liu2017retrosynthetic,karpov2019transformer,kim2021valid,sacha2021molecule,wang2021retroprime,zhong2022root,zhong2023retrosynthesis}, which indicates the proportion of ground truth reactants that appear (exactly matched) among the top-\textit{K} confident predicted reactants, to evaluate the performance on each client. For comprehensive evaluation, we additionally report MaxFrag accuracy~\cite{tetko2020state} which focuses on main compound transformations, and RoundTrip accuracy~\cite{chen2021deep,wan2022retroformer} which evaluates whether the predicted reactants can indeed synthesize the target product.

\bmhead{Data availability}

All chemical reaction data used in this work are publicly available. The USPTO-50K dataset is available in the Zenodo repository under accession code 10.5281/zenodo.15679564 (\href{https://zenodo.org/records/15679564}{https://zenodo.org/records/15679564})~\cite{Chen2025CKIF}.
The USPTO-MIT dataset is available at (\href{https://github.com/wengong-jin/nips17-rexgen/blob/master/USPTO/data.zip}{https://github.com/wengong-jin/nips17-rexgen/blob/master/USPTO/data.zip}).
The USPTO 1k TPL dataset is available at (\href{https://ibm.ent.box.com/v/MappingChemicalReactions}{https://ibm.ent.box.com/v/MappingChemicalReactions}). No new primary datasets were generated in this study. Source data are provided with this paper.

\bmhead{Code availability}

The source code and implementation details that support the findings of this study are publicly available under the MIT License at \href{https://github.com/guikunchen/CKIF}{https://github.com/guikunchen/CKIF}~\cite{Chen2025CKIF}.

\bibliography{sn-bibliography}

\bmhead{Competing Interests Statement}
The authors declare no competing interests.

\clearpage

\begin{table}[t]
    \setlength{\tabcolsep}{4.4pt}
    \centering
    \caption{Top-\textit{K} retrosynthesis accuracy (\%) on the USPTO-50K dataset. Note that Centrally Trained gathers all reaction data and the comparison is unfair. Bold values indicate the best results among Locally Trained, FedAvg~\cite{mcmahan2017communication}, and CKIF (chemical knowledge-informed framework). MF and RT denote MaxFrag accuracy and RoundTrip accuracy, respectively.}
    \label{tab:t1}
    \begin{tabular}{llcccc|cccc|c}
    \toprule
    Client       & Method                     & \textit{K} = 1 & 3    & 5    & 10   & 1 (MF)   & 3  & 5 & 10 & 1 (RT) \\ \midrule
    $C_1$ & Locally Trained   & 41.9 & 57.1 & 65.0 & 69.8 & 56.8 & 69.7 & 73.4 & 75.5 & 51.0 \\
                   & Centrally Trained & 40.1 & 58.8 & 69.1 & 73.9 & 51.0 & 68.2 & 76.1 & 79.4 & 59.2 \\
                   & FedAvg~\cite{mcmahan2017communication}  & 15.0 & 30.9 & 37.2 & 40.8 & 19.2 & 38.3 & 45.0 & 48.7 & 30.3 \\
                   & CKIF (Ours)     & \textbf{43.9} & \textbf{60.2} & \textbf{67.1} & \textbf{70.3} & 56.5 & \textbf{71.1} & \textbf{76.3} & \textbf{78.0} & \textbf{51.2} \\
    \midrule
    $C_2$ & Locally Trained   & 4.1 & 8.6 & 9.2 & 11.1 & 13.8 & 18.2 & 19.4 & 21.5 & 6.9 \\
                   & Centrally Trained & 19.0 & 28.6 & 33.7 & 37.0 & 27.2 & 40.4 & 43.9 & 47.8 & 52.6 \\
                   & FedAvg~\cite{mcmahan2017communication}  & 0.0 & 0.4 & 0.9 & 1.2 & 0.2 & 1.1 & 2.1 & 2.8 & 16.4 \\
                   & CKIF (Ours)     & \textbf{23.6} & \textbf{33.3} & \textbf{37.6} & \textbf{40.0} & \textbf{36.7} & \textbf{46.9} & \textbf{50.6} & \textbf{52.7} & \textbf{41.3} \\
    \midrule
    $C_3$ & Locally Trained   & 28.0 & 41.6 & 46.1 & 47.7 & 29.9 & 43.2 & 47.3 & 48.6 & 30.7 \\
                   & Centrally Trained & 23.6 & 39.5 & 45.3 & 49.0 & 24.0 & 40.6 & 46.8 & 50.5 & 46.9 \\
                   & FedAvg~\cite{mcmahan2017communication}  & 12.7 & 23.2 & 27.2 & 28.7 & 12.8 & 23.7 & 27.5 & 29.1 & 27.8 \\
                   & CKIF (Ours)     & \textbf{30.9} & \textbf{47.4} & \textbf{52.9} & \textbf{55.5} & \textbf{32.3} & \textbf{48.7} & \textbf{54.0} & \textbf{56.6} & \textbf{35.9} \\
    \midrule
    $C_4$ & Locally Trained   & 37.1 & 49.9 & 54.9 & 56.9 & 39.1 & 51.7 & 57.0 & 58.9 & 40.2 \\
                   & Centrally Trained & 39.3 & 58.3 & 65.5 & 70.6 & 42.3 & 61.0 & 68.0 & 72.0 & 50.7 \\
                   & FedAvg~\cite{mcmahan2017communication}  & 12.3 & 29.0 & 35.3 & 37.9 & 13.2 & 30.8 & 37.2 & 40.5 & 27.2 \\
                   & CKIF (Ours)     & \textbf{39.4} & \textbf{55.3} & \textbf{60.9} & \textbf{63.2} & \textbf{41.1} & \textbf{57.1} & \textbf{62.5} & \textbf{64.4} & \textbf{43.7} \\
    \botrule
    \end{tabular}
\end{table}

\begin{table}[t]
    \setlength{\tabcolsep}{4.4pt}
    \centering
    \caption{Top-\textit{K} retrosynthesis accuracy (\%) on the USPTO-50K and USPTO-MIT dataset. Note that Centrally Trained gathers all reaction data and the comparison is unfair. Bold values indicate the best results among Locally Trained, FedAvg~\cite{mcmahan2017communication}, and CKIF (chemical knowledge-informed framework). Here MF and RT denote MaxFrag accuracy and RoundTrip accuracy, respectively. The low RoundTrip accuracy for $C_6$ can be attributed to the used pre-trained forward synthesis model not having encountered $C_6$'s reactions during training.}
    \label{tab:t2}
    \begin{tabular}{llcccc|cccc|c}
    \toprule
    Client       & Method                     & \textit{K} = 1 & 3    & 5    & 10   & 1 (MF)   & 3  & 5 & 10 & 1 (RT) \\ \midrule
    $C_1$ & Locally Trained   & 41.9 & 57.1 & 65.0 & 69.8 & 56.8 & 69.7 & 73.4 & 75.5 & 51.0  \\
                   & Centrally Trained & 30.8 & 47.5 & 56.1 & 65.1 & 39.5 & 57.3 & 66.0 & 74.0 & 57.1  \\
                   & FedAvg~\cite{mcmahan2017communication}  & 14.1 & 31.0 & 39.8 & 45.4 & 18.6 & 40.0 & 49.8 & 55.5 & 31.9  \\
                   & CKIF (Ours)     & \textbf{44.2} & \textbf{64.5} & \textbf{72.4} & \textbf{77.0} & \textbf{58.7} & \textbf{76.1} & \textbf{80.7} & \textbf{82.8} & \textbf{57.1}  \\
    \midrule
    $C_2$ & Locally Trained   & 4.1 & 8.6 & 9.2 & 11.1 & 13.8 & 18.2 & 19.4 & 21.5 & 6.9  \\
                   & Centrally Trained & 18.0 & 33.5 & 39.0 & 45.5 & 30.2 & 48.1 & 54.5 & 60.0 & 54.9  \\
                   & FedAvg~\cite{mcmahan2017communication}  & 15.5 & 39.8 & 48.8 & 56.6 & 18.8 & 48.8 & 57.1 & 64.3 & 19.9  \\
                   & CKIF (Ours)     & \textbf{30.7} & \textbf{43.6} & \textbf{48.9} & \textbf{52.6} & \textbf{42.2} & \textbf{57.0} & \textbf{61.6} & \textbf{64.0} & \textbf{51.1}  \\
    \midrule
    $C_3$ & Locally Trained   & 28.0 & 41.6 & 46.1 & 47.7 & 29.9 & 43.2 & 47.3 & 48.6 & 30.7  \\
                   & Centrally Trained & 18.0 & 36.0 & 43.3 & 48.4 & 18.6 & 36.6 & 44.4 & 49.4 & 52.3  \\
                   & FedAvg~\cite{mcmahan2017communication}  & 1.6 & 5.3 & 7.1 & 9.5 & 2.6 & 7.9 & 10.4 & 13.9 & 37.2  \\
                   & CKIF (Ours)     & \textbf{36.9} & \textbf{56.5} & \textbf{62.4} & \textbf{64.8} & \textbf{38.4} & \textbf{57.6} & \textbf{63.3} & \textbf{65.8} & \textbf{40.2}  \\
    \midrule
    $C_4$ & Locally Trained   & 37.1 & 49.9 & 54.9 & 56.9 & 39.1 & 51.7 & 57.0 & 58.9 & 30.7  \\
                   & Centrally Trained & 20.4 & 33.0 & 40.3 & 46.6 & 23.7 & 37.4 & 45.0 & 49.5 & 52.3  \\
                   & FedAvg~\cite{mcmahan2017communication}  & 15.3 & 32.5 & 38.7 & 42.0 & 15.3 & 32.7 & 38.9 & 42.5  & 37.2 \\
                   & CKIF (Ours)     & \textbf{40.9} & \textbf{58.7} & \textbf{63.4} & \textbf{67.1} & \textbf{42.2} & \textbf{60.7} & \textbf{64.8} & \textbf{68.3} & \textbf{47.7}  \\
    \midrule

    $C_5$ & Locally Trained   & 36.5 & 51.8 & 59.0 & 63.9 & 50.1 & 63.9 & 68.7 & 71.4 & 55.5  \\
                   & Centrally Trained & 23.2 & 36.0 & 42.8 & 48.7 & 30.9 & 45.5 & 52.8 & 59.3 & 53.4  \\
                   & FedAvg~\cite{mcmahan2017communication}  & 13.0 & 29.0 & 34.8 & 41.4 & 14.5 & 31.8 & 38.2 & 44.5 & 35.8  \\
                   & CKIF (Ours)   & 34.6 & \textbf{53.2} & \textbf{62.6} & \textbf{67.5} & 48.5 & \textbf{65.3} & \textbf{72.0} & \textbf{74.7} & \textbf{56.9}   \\
    \midrule
    $C_6$ & Locally Trained   & 11.7 & 17.8 & 20.5 & 22.5 & 19.3 & 27.2 & 29.6 & 31.6 & 0.0  \\
                   & Centrally Trained & 18.7 & 31.9 & 36.7 & 42.0 & 26.6 & 40.2 & 45.1 & 50.1 & 0.6  \\
                   & FedAvg~\cite{mcmahan2017communication}  & 10.5 & 24.7 & 30.5 & 34.8 & 13.0 & 31.6 & 37.6 & 42.4 & 0.6  \\
                   & CKIF (Ours)     & \textbf{21.7} & \textbf{33.3} & \textbf{38.2} & \textbf{42.1} & \textbf{28.9} & \textbf{40.3} & \textbf{45.2} & \textbf{48.8} & 0.6   \\
    \midrule
    $C_7$ & Locally Trained   & 20.6 & 29.0 & 32.6 & 34.5 & 27.5 & 36.2 & 39.4 & 40.9 & 22.9  \\
                   & Centrally Trained & 21.8 & 34.4 & 39.8 & 43.7 & 27.2 & 41.6 & 46.7 & 50.6 & 45.6  \\
                   & FedAvg~\cite{mcmahan2017communication}  & 2.3 & 7.6 & 10.7 & 12.9 & 3.7 & 9.6 & 13.6 & 16.7 & 27.1  \\
                   & CKIF (Ours)     & \textbf{35.8} & \textbf{49.4} & \textbf{54.2} & \textbf{57.7} & \textbf{44.5} & \textbf{56.4} & \textbf{60.9} & \textbf{63.5} & \textbf{40.6}   \\
    \midrule
    $C_8$ & Locally Trained   & 18.3 & 25.7 & 29.9 & 31.5 & 25.8 & 34.1 & 37.0 & 38.5 & 30.4  \\
                   & Centrally Trained & 27.7 & 41.6 & 47.7 & 51.8 & 33.2 & 49.2 & 55.1 & 58.7 & 69.6  \\
                   & FedAvg~\cite{mcmahan2017communication}  & 5.0 & 7.0 & 8.4 & 9.2 & 6.8 & 12.0 & 15.1 & 17.2 & 42.1  \\
                   & CKIF (Ours)     & \textbf{25.9} & \textbf{38.0} & \textbf{43.2} & \textbf{46.4} & \textbf{32.0} & \textbf{45.8} & \textbf{50.1} & \textbf{52.5} & \textbf{56.2}   \\
    
    \botrule
    \end{tabular}
\end{table}

\begin{table}[b]
    \setlength{\tabcolsep}{4.4pt}
    \centering
    \caption{Top-\textit{K} retrosynthesis accuracy (\%) on the USPTO 1k TPL dataset. Note that Centrally Trained gathers all reaction data and the comparison is unfair. Bold values indicate the best results among Locally Trained, FedAvg~\cite{mcmahan2017communication}, and CKIF (chemical knowledge-informed framework). Here MF and RT denote MaxFrag accuracy and RoundTrip accuracy, respectively.}
    \label{tab:t3}
    \begin{tabular}{llcccc|cccc|c}
    \toprule
    Client       & Method                     & \textit{K} = 1 & 3    & 5    & 10   & 1 (MF)   & 3  & 5 & 10 & 1 (RT) \\ \midrule
    $C_9$ & Locally Trained  & 53.8 & 67.4 & 73.3 & 76.6 & 63.1 & 74.3 & 78.5 & 80.0 & 53.1  \\
                   & Centrally Trained & 50.8 & 71.0 & 80.1 & 86.3 & 58.2 & 77.6 & 84.7 & 88.9 & 69.5  \\
                   & FedAvg~\cite{mcmahan2017communication}  & 44.5 & 64.6 & 72.3 & 77.1 & 49.7 & 70.1 & 76.7 & 81.0 & 61.7  \\
                   & CKIF (Ours)     & \textbf{63.5} & \textbf{77.6} & \textbf{85.2} & \textbf{88.7} & \textbf{70.3} & \textbf{83.7} & \textbf{88.7} & \textbf{90.6} & \textbf{63.5}   \\
    \midrule
    $C_{10}$ & Locally Trained   & 60.4 & 72.2 & 76.9 & 79.4 & 66.4 & 76.9 & 80.1 & 81.7 & 53.3  \\
                   & Centrally Trained & 31.9 & 54.1 & 64.3 & 70.8 & 44.2 & 65.7 & 74.9 & 80.1 & 67.0  \\
                   & FedAvg~\cite{mcmahan2017communication}  & 38.0 & 57.5 & 66.4 & 72.0 & 45.7 & 65.5 & 74.5 & 80.5 & 59.2  \\
                   & CKIF (Ours)     & \textbf{66.6} & \textbf{80.2} & \textbf{86.3} & \textbf{90.3} & \textbf{71.4} & \textbf{85.0} & \textbf{89.0} & \textbf{91.4} & \textbf{63.5}   \\
    \midrule
    $C_{11}$ & Locally Trained   & 55.5 & 65.9 & 70.7 & 73.5 & 61.7 & 71.4 & 74.5 & 75.7 & 48.2  \\
                   & Centrally Trained & 40.7 & 61.0 & 70.3 & 76.2 & 50.3 & 69.3 & 75.9 & 79.9 & 66.3  \\
                   & FedAvg~\cite{mcmahan2017communication}  & 42.2 & 59.7 & 67.8 & 74.6 & 48.8 & 65.7 & 71.9 & 77.5 & 60.2  \\
                   & CKIF (Ours)     & \textbf{62.0} & \textbf{76.1} & \textbf{84.5} & \textbf{87.9} & \textbf{69.8} & \textbf{82.4} & \textbf{87.5} & \textbf{89.2} & \textbf{63.3}   \\
    \midrule
    $C_{12}$ & Locally Trained   & 55.8 & 69.2 & 73.5 & 76.0 & 62.4 & 73.8 & 77.5 & 79.3 & 51.1  \\
                   & Centrally Trained & 33.9 & 54.6 & 65.6 & 72.7 & 43.7 & 64.2 & 72.7 & 77.4 & 63.0  \\
                   & FedAvg~\cite{mcmahan2017communication}  & 37.4 & 55.2 & 62.6 & 68.3 & 45.8 & 63.8 & 69.8 & 73.9 & 58.8  \\
                   & CKIF (Ours)     & \textbf{58.7} & \textbf{75.3} & \textbf{83.9} & \textbf{87.7} & \textbf{66.1} & \textbf{81.4} & \textbf{87.4} & \textbf{89.3} & \textbf{61.0}   \\
    \midrule

    $C_{13}$ & Locally Trained   & 56.4 & 69.7 & 75.9 & 79.0 & 63.5 & 75.3 & 78.9 & 81.0 & 51.9  \\
                   & Centrally Trained & 40.7 & 62.1 & 72.2 & 77.8 & 49.1 & 68.8 & 76.7 & 80.7 & 66.5  \\
                   & FedAvg~\cite{mcmahan2017communication}  & 39.7 & 60.0 & 68.5 & 73.6 & 45.2 & 64.4 & 71.8 & 76.0 & 61.2  \\
                   & CKIF (Ours)   & \textbf{64.2} & \textbf{78.6} & \textbf{86.1} & \textbf{89.3} & \textbf{71.5} & \textbf{83.9} & \textbf{88.7} & \textbf{90.6} & \textbf{62.1}   \\
    \midrule
    $C_{14}$ & Locally Trained   & 53.6 & 66.1 & 71.2 & 74.8 & 62.1 & 71.6 & 75.6 & 77.6 & 45.4  \\
                   & Centrally Trained & 54.1 & 73.1 & 83.5 & 89.3 & 63.2 & 81.0 & 87.8 & 91.5 & 61.5  \\
                   & FedAvg~\cite{mcmahan2017communication}  & 35.3 & 54.8 & 63.5 & 69.3 & 43.4 & 62.8 & 70.7 & 75.3 & 53.0  \\
                   & CKIF (Ours)     & \textbf{61.5} & \textbf{76.7} & \textbf{83.1} & \textbf{87.5} & \textbf{69.5} & \textbf{82.1} & \textbf{87.1} & \textbf{89.1} & \textbf{55.0}   \\
    \midrule
    $C_{15}$ & Locally Trained   & 55.8 & 69.2 & 73.9 & 76.5 & 63.4 & 74.6 & 78.1 & 79.5 & 52.2  \\
                   & Centrally Trained & 59.0 & 77.4 & 89.9 & 96.0 & 68.2 & 87.0 & 94.5 & 96.4 & 72.3  \\
                   & FedAvg~\cite{mcmahan2017communication}  & 40.9 & 59.8 & 69.1 & 74.5 & 48.1 & 66.7 & 75.1 & 78.8 & 61.8  \\
                   & CKIF (Ours)     & \textbf{61.2} & \textbf{77.1} & \textbf{87.3} & \textbf{91.1} & \textbf{70.7} & \textbf{84.2} & \textbf{90.2} & \textbf{92.2} & \textbf{64.3}   \\
    \midrule
    $C_{16}$ & Locally Trained   & 54.0 & 67.3 & 74.8 & 78.5 & 63.8 & 75.8 & 80.0 & 82.3 & 49.9  \\
                   & Centrally Trained & 56.5 & 75.9 & 86.0 & 91.8 & 66.2 & 83.7 & 90.3 & 93.1 & 67.4  \\
                   & FedAvg~\cite{mcmahan2017communication}  & 39.0 & 58.8 & 67.8 & 73.3 & 47.3 & 66.0 & 73.5 & 78.4 & 56.4  \\
                   & CKIF (Ours)     & \textbf{62.6} & \textbf{76.9} & \textbf{86.2} & \textbf{89.8} & \textbf{70.2} & \textbf{84.0} & \textbf{89.8} & \textbf{91.4} & \textbf{61.0} \\
    
    \botrule
    \end{tabular}
\end{table}

\clearpage

\end{document}